\documentclass[10pt,twocolumn,letterpaper]{article}

\usepackage[pagenumbers]{cvpr} 

\usepackage{graphicx}
\usepackage{amsmath}
\usepackage{amssymb}
\usepackage{booktabs}
\usepackage{threeparttable}
\usepackage{multirow}
\usepackage{amsmath, amssymb, amsthm}
\usepackage{xcolor}
\usepackage[ruled,vlined]{algorithm2e}
\usepackage{float}
 
\usepackage{booktabs}   
\usepackage{amsfonts}       
\usepackage{nicefrac}       
\usepackage{microtype}      
\usepackage{bm}

\usepackage[pagebackref,breaklinks,colorlinks]{hyperref}

\usepackage[capitalize]{cleveref}
\crefname{section}{Sec.}{Secs.}
\Crefname{section}{Section}{Sections}
\Crefname{table}{Table}{Tables}
\crefname{table}{Tab.}{Tabs.}

\def\cvprPaperID{6278} 
\def\confName{CVPR}
\def\confYear{2022}

\begin{document}

\title{WPNAS: Neural Architecture Search by jointly using Weight Sharing and Predictor}

\author{\hspace{-2ex} Ke Lin$^{1}$\thanks{Work done during the author works at Samsung Research China, Beijing (SRC-B).} \quad Yong A$^{2}$ \quad Zhuoxin Gan$^{2}$ \quad Yingying Jiang$^{2}$  \\[2mm]
\hspace{-2ex}$^1$Huawei  \quad $^2$Samsung Research China, Beijing (SRC-B)}
\maketitle

\begin{abstract}
   Weight sharing based and predictor based methods are two major types of fast neural architecture search methods. In this paper, we propose to jointly use weight sharing and predictor in a unified framework. First, we construct a SuperNet in a weight-sharing way and probabilisticly sample architectures from the SuperNet. To increase the correctness of the evaluation of architectures, besides direct evaluation using the inherited weights, we further apply a few-shot predictor to assess the architecture on the other hand. The final evaluation of the architecture is the combination of direct evaluation, the prediction from the predictor and the cost of the architecture. We regard the evaluation as a reward and apply a self-critical policy gradient approach to update the architecture probabilities. To further reduce the side effects of weight sharing, we propose a weakly weight sharing method by introducing another HyperNet. We conduct experiments on datasets including CIFAR-10, CIFAR-100 and ImageNet under NATS-Bench, DARTS and MobileNet search space. The proposed WPNAS method achieves state-of-the-art performance on these datasets.
\end{abstract}

\section{Introduction}
\label{sec:intro}

Neural Architecture Search (NAS) is proposed to make the computer automatically search for an optimal neural network architecture, whose performance can surpass the manually designed state-of-the-art neural network architectures and has similar or less computational complexity (e.g, latency, FLOPs) \cite{zoph}. Early NAS methods usually sample architectures from a pre-defined search space, obtain the performance of the architecture after training from scratch, and update the search algorithm through reinforcement learning \cite{zoph, tan2019mnasnet} or evolution algorithm \cite{real2019regularized, yang2020cars} until one or a cluster of optimal architectures are obtained. Although these methods have achieved amazing performances, they often need thousands of GPU-day to run the searching algorithm. Weight-sharing \cite{pham2018efficient, liu2018darts} and predictor \cite{wen2020predictor, ning2020gates} are two most commonly used methods to solve the above problem of excessive resource consumption.

Weight sharing refers to building a SuperNet and all sub-architectures in the search space can inherit weights from the SuperNet. Therefore, the amount of parameters to be trained in the search process and the search time can be greatly reduced. Weight sharing NAS algorithms can be divided into two categories: i) Gradient based which parametrize architecture parameter as part of the SuperNet and update the parameter of the SuperNet (architecture parameters and weight parameters) in a differentiable way \cite{liu2018darts, xu2019pc, chu2020fairdarts, xue2021idarts}, ii) Sampling based algorithms including uniform sampling \cite{guo2019single, chu2021fairnas} and learning based sampling \cite{casale2019PARSEC, chen2020adc, yan2021fpnas}. Because all sub-architectures are deeply coupled with each other, the evaluation of the performance of each architecture have large gaps between those architectures obtained by stand-alone training \cite{chu2021fairnas}. Therefore, in order to improve the ranking consistency between the architectures of weight sharing and the architectures obtained by stand-alone training, there are still two major problems to be solved: i) how to improve the correctness of the evaluation of each architecture and ii) is there any way to reduce the degree of weight sharing.

Predictor based NAS algorithms \cite{wen2020predictor, ning2020gates} are proposed thanks to the introduction of several excellent benchmarking datasets such as NAS-Bench-101 \cite{ying2019bench101}, NAS-Bench-201 \cite{dong2020bench201} and NATS-Bench \cite{dong2020natsbench}. Predictor based NAS collects several architecture-accuracy pairs as training data in advance, and uses MLP \cite{wen2020predictor} or GCN \cite{ning2020gates} as predictor to train the direct mapping from architecture to accuracy. However, collecting architecture-accuracy pairs is a resource consuming issue, because it needs to fully train each architecture from scratch of the training pairs. How to improve the training efficiency of the predictor to reduce the dependence on training data is also an important question to be answered.

\begin{figure*}[t]
	
	\centering
	\includegraphics[width=0.9\textwidth]{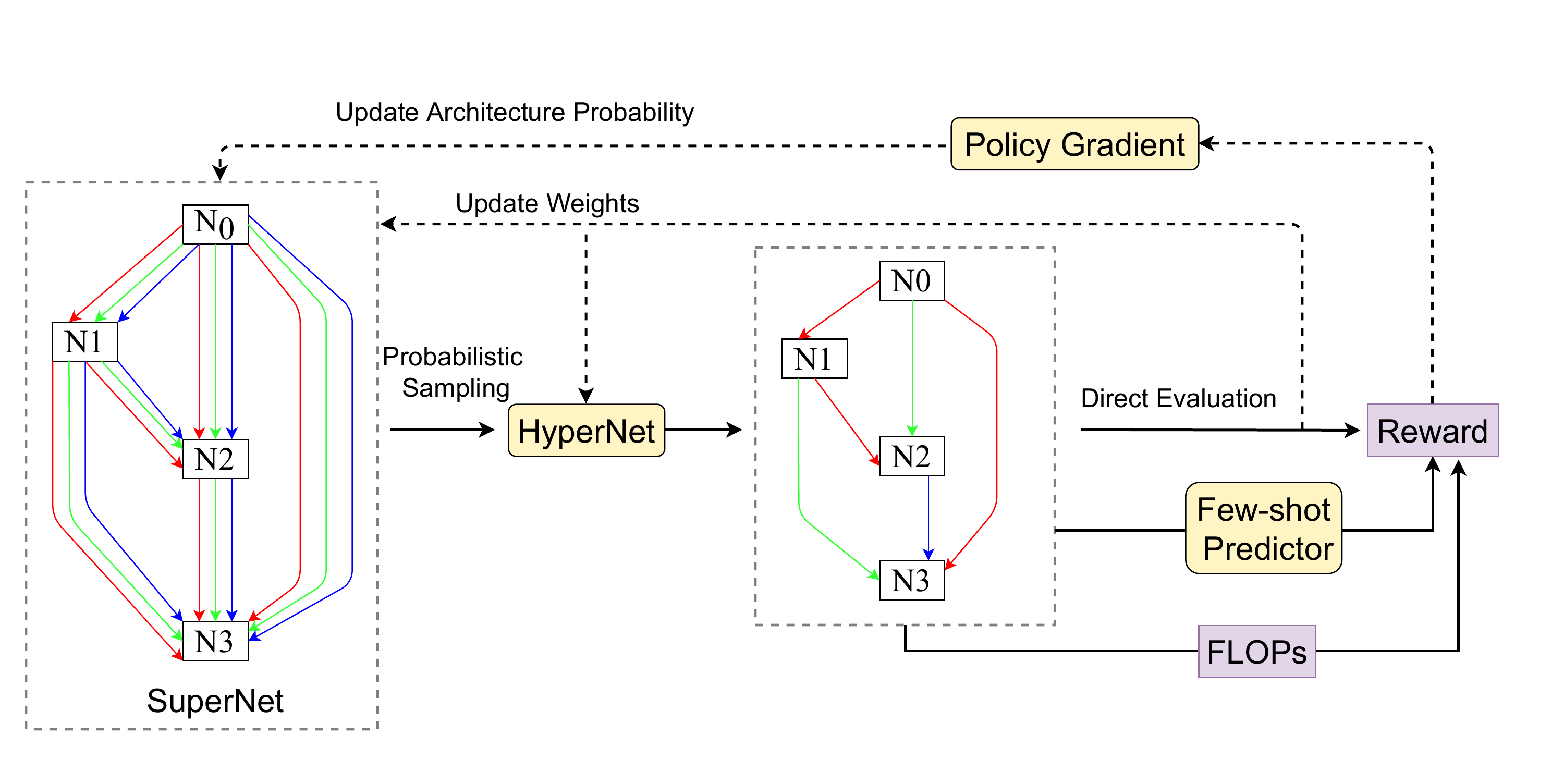}

	\caption{Overview of the proposed WPNAS. We apply a probabilistic sampling strategy to sample architectures from the SuperNet and use a HyperNet to alleviate the influence of weight sharing. The evaluation of the sampled architecture comes from three parts: direct evaluation, prediction from the few-shot predictor and the FLOPs of the architecture. Then we regard the evaluation of the architecture as reward and use a policy gradient algorithm to update architecture probabilities. The combined reward contains information from both SuperNet and predictor, thus may lead better evaluation. Few-shot learning is proposed to reduce the demand for training data of the predictor.}
	\label{supp_f3}
\end{figure*}

In this paper, we attempt to answer the above questions using a Weight-sharing and Predictor based NAS (WPNAS) method. First, we construct a SuperNet in a weight-sharing way and probabilistic sampling architectures from the SuperNet with inherited weights. To increase the correctness of the evaluation of architectures, besides direct evaluation, we further apply a predictor to predict the performance the architecture on the other hand. Direct evaluation from SuperNet, the prediction from the predictor and the cost of the architecture are regarded as the reward and we apply a self-critical policy gradient approach to update the architecture probabilities. To increase the training efficiency of the predictor, we propose a few-shot learning based strategy to train the predictor. Besides, to further reduce the negative effects of weight sharing, we propose a weakly weight sharing method. Namely, we feed the sampled architecture into a HyperNet and generate different offset-weights, the final weights is the product of the offset-weights and weights from SuperNet. By this way, the degree of weight sharing is reduced.

In summary, the contribution of this work contains four major parts:
\begin{itemize}
	\item 
	We jointly use weight-sharing and predictor and use a self-critical policy gradient algorithm with probabilistic sampling to update architecture parameters.  
	\item We propose a few-shot learning based predictor to increase the training efficiency of the predictor.
	\item HyperNet based weakly weight sharing strategy is proposed to alleviate the influence of weight sharing.
	\item We conduct comprehensive experiments on CIFAR and ImageNet datasets under several search space and the results demonstrate the superiority of the proposed method.
\end{itemize}

\begin{figure*}[t]
	
	\centering
	\includegraphics[width=1\textwidth]{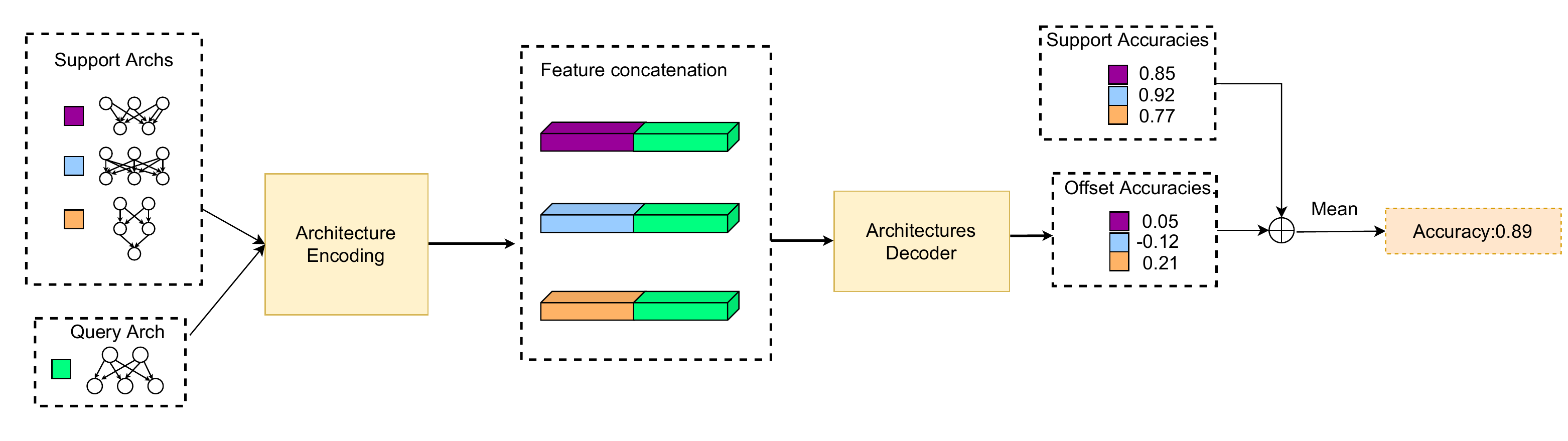}

	\caption{Few-shot learning based predictor.}
	\label{supp_f3}
\end{figure*}

\section{Related Work}
\label{sec:intro}

\noindent\textbf{Weight sharing based NAS}.
ENAS \cite{pham2018efficient} is the first NAS paper to apply weight sharing to address the high searching cost problem. ENAS \cite{pham2018efficient} applies reinforcement learning to train a controller to sample architectures and it does not need to train each sampled architecture from scratch, but directly inherit parameters from the Supernet, so the architecture search time can be reduced from thousands of GPU-day to only less than a single GPU-day. DARTS \cite{liu2018darts} uses a continuous relaxation on the categorical choices of inputs and operations. The weight parameters and architecture parameters can be updated based on the gradient, and the search time can be further reduced. DARTS has received a lot of attention and follow-up research because of its high simplicity and effectiveness \cite{xu2019pc, chen2019progressive, chu2020fairdarts, xue2021idarts, chu2021darts-}. The above gradient based methods still face two important problems: high GPU memory occupation and lack of theoretical support for discretization. \cite{dong2019searching} and \cite{chang2019data} propose to apply Gumbel-Softmax trick to reduce the GPU memory occupation. Sampling based methods are proposed to solve the above two problems at the same time. SPOS \cite{guo2019single} proposes a two-stage algorithm that abandoning the architecture parameters, using only uniform sampling to train the Supernet, and applying evolution algorithm to search optimal architectures on the trained Supernet which can also achieve excellent performance. Some follow-up studies also improved this unbiased sampling method from many aspects \cite{chu2021fairnas, huang2021generator, you2020greedy}. Learning based sampling can favor the most promising architectures and learn the optimal sampling probabilities. PARSEC \cite{casale2019PARSEC} and FP-NAS  \cite{yan2021fpnas} apply probabilistic sampling and use an empirical Bayes Monte Carlo procedure to optimize architecture parameters. MCT-NAS \cite{su2021mctnas} challenges the operation independency assumption and applies a Monte Carlo tree search strategy which captures the dependency among layers and operations. Our proposed method falls into the probabilistic sampling based NAS method category, but we use a different self-critical policy gradient way to optimize the architecture probabilities.

\noindent\textbf{Predictor based NAS}.
Predictor based NAS methods are proposed recently thanks to the introduction of some benchmark dataset. \cite{wen2020predictor} first apply a predictor to regress the validation accuracy. They train several architectures to obtain architecture-accuracy pairs as training data to train the regression model to predict the validation accuracy directly. The architecture with the best predicted accuracy is trained from scratch for the final deployment. \cite{ning2020gates, dudziak2020gcn} applies Graph convolutional networks (GCN) to embed the architectures to better capture the topology information.  \cite{xu2021renas, chen2021comparator} considers the ranking property between architectures and apply rank loss to optimize the predictor. \cite{tang2020semi} using semi-supervised learning to reduce the dependency on training samples. In this paper, we reduce the demand for training samples from another perspective by using few-shot learning.

\section{WPNAS}
\label{sec:intro}

In this section, we will present our WPNAS algorithm. In Section 3.1, we introduce the probabilistic sampling method; in Section 3.2, we describe the novel few-shot predictor; in section 3.3, we introduce the weakly weight sharing approach; in Section 3.4, we summary the optimization algorithm.

\begin{figure*}[t]
	
	\centering
	\includegraphics[width=1\textwidth]{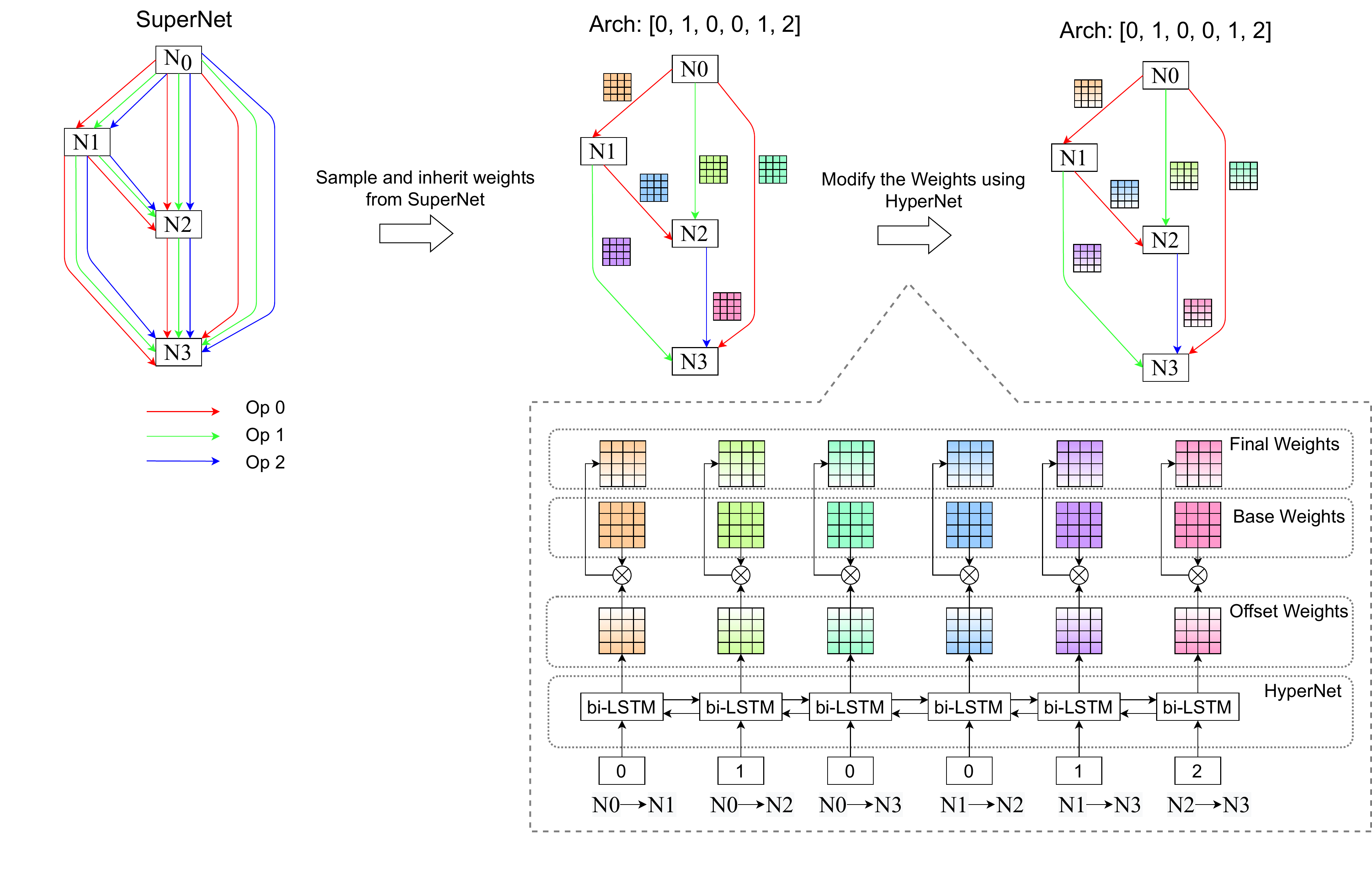}

	\caption{Weakly weight sharing by HyperNet. Different architecture can generate different offset weights, thus obtaining different final weights.}
	\label{supp_f3}
\end{figure*}

\subsection{Probabilistic Sampling}

Suppose the SuperNet with model weights $\bm{\omega}$ with a macro search space has $L$ layers, each layer have $K$ candidate operations $\{o_1,...,o_K\}$. An individual architecture $\mathbf{A}$ from the SuperNet can be denoted as $\mathbf{A} = (A^1,...,A^L), A^l\in\{o_1,...,o_K\}$. Like \cite{casale2019PARSEC, yan2021fpnas}, we also introduce architecture parameters (i.e. architecture probabilities) $\bm{\alpha}$ and a prior distribution $p(\mathbf{A}|\bm{\alpha})$. Each architecture $\mathbf{A}$ is sampled based on $p(\mathbf{A}|\bm{\alpha})$, thus the task of architecture search is to learn the  distribution $p(\mathbf{A}|\bm{\alpha})$ which prefers better architectures. Suppose the probabilities at different layers are independent to each other, the probability of architecture $\mathbf{A}$ being sampled can be calculated as below:

\begin{equation}
p(\mathbf{A}|\bm{\alpha}) = \prod_lp(A^l|\alpha^l)
\end{equation}

\subsection{Few-shot Predictor}

Traditional accuracy predictor of NAS always needs thousands of training data pairs which require tremendous computing resources to obtain. To address this problem, we borrow the idea from few-shot learning on image classification for more efficient predictor training. Specifically, we construct a simple yet effective Relation Network \cite{sung2018relation}-like predictor by learning to compare query architectures and support architectures. The predictor has two major modules: architecture encoding module and architecture decoder module (see Figure 2.). The architecture encoding module embeds the support architectures and query architecture into the hidden representations with the same dimension. The embeddings from support set and query architecture are concatenated and are fed to the architecture decoder module to learn the accuracy offset between each architecture in the support set and the query architecture. The encoding and decoding modules can be MLP based, RNN based, Transformer based or GCN based. During training, the support set and the query set are randomly chosen from the training data. Once trained, the few-shot predictor is able to predict new architectures by computing relation offsets between the query architecture and the few examples of training set. For simplicity, we denote the output of the few-shot predictor as $FSP(\bm{A})$. In the following section, we will demonstrate the superiority of the proposed few-shot predictor than supervised learning based predictor.

\subsection{Weakly weight sharing}

Although weight sharing brings benefits such as reducing search consumption, it also leads to the fatal problem of the ranking gap between architectures trained by weight sharing and stand-alone training. To relieve this problem, we apply a weakly weight sharing strategy by introducing a HyperNet (see Figure 3.). Specifically, for a sampled architecture $\mathbf{A} = (A^1,...,A^L)$, the weights of the architecture can be inherit from the SuperNet, and we call these weights as base weights $\bm{\omega}_b = [\bm{\omega}_b^1,..., \bm{\omega}_b^L]$. A bidirectional LSTM (bi-LSTM) with $L$ time steps is used as the HyperNet to generate several weights for each edge (or layer) given the architecture, and these weights are called offset weights $\bm{\omega}_o = [\bm{\omega}_o^1,..., \bm{\omega}_o^L]$. The final weights for evaluation and back-propagation are the product of the based weights and offset weights:

\begin{equation}
\bm{\omega} = [\bm{\omega}^1,..., \bm{\omega}^L], \bm{\omega}^l = \bm{\omega}_b^l\bm{\omega}_o^l
\end{equation}

Suppose two architectures choose the same operation in a certain layer. If there is no weakly weight sharing, the two architectures have exactly the same weights on that layer. After the introduction of HyperNet, different architectures will generate different offset weights, if the two architectures have the same operation in a certain layer, their weights at that layer will be different. Thus, the introduction of HyperNet will reduce the degree of weight sharing among architectures, which may alleviate the problem of the ranking gap. The weights of HyperNet can be fixed or updated simultaneously with the SuperNet.

The maximum kernel size we used in this study is 5, so the dimension of $\bm{\omega}_o^l$ is $5\times5=25$. For operations with smaller kernel size such as $3\times3$, the weights of the top left corner are cropped and used.

\subsection{Optimization}

Given a classification task with input $\mathbf{X}$ and targets $\mathbf{y}$, the goal of optimization can be present as:

\begin{equation}
p(\mathbf{y}|\mathbf{X},\bm{\alpha},\bm{\omega}) = \int p(\mathbf{y}|\mathbf{X},\bm{A},\bm{\omega})p(\mathbf{A}|\bm{\alpha})d\mathbf{A}
\end{equation}

We can simplify the optimization through Monte Carlo sampling and policy gradient algorithm. For $\bm{\omega}$, the optimization is straight forward. For a sampled architecture $\bm{\widetilde{A}}$ with weights $\bm{\widetilde{\omega}}$, we can directly optimize $\bm{\widetilde{\omega}}$ by $\nabla_{\bm{\widetilde{\omega}}}\log p(\mathbf{y}|\mathbf{X},\bm{\widetilde{A}},\bm{\widetilde{\omega}})$.

For $\bm{\alpha}$, the gradient of loss can be estimated by policy gradient:
\begin{equation}
\label{111}
\nabla_{\bm{\alpha}}L = -r\nabla_{\bm{\alpha}}\log p(\bm{\widetilde{A}}|\bm{\alpha})
\end{equation}

\noindent where $r$ is the reward function. The reward function contains three parts: 1) direct evaluation, i.e. the negative cross-entrophy loss of the architecture $\log p(\mathbf{y}|\mathbf{X},\bm{\widetilde{A}},\bm{\widetilde{\omega}})$ ; 2) the output of the few-shot predictor $FSP(\bm{\widetilde{A}})$ ; 3) and the negative FLOPs of the sampled architecture $-FLOPs(\bm{\widetilde{A}}, \bm{X})$. Thus, we get:

\begin{equation}
\small
\begin{array}{l}
\nabla_{\bm{\alpha}}L = 
-[\log p(\mathbf{y}|\mathbf{X},\bm{\widetilde{A}},\bm{\widetilde{\omega}}) + \beta_1 FSP(\bm{\widetilde{A}}) - \beta_2 FLOPs(\bm{\widetilde{A}}, \bm{X})] \\ 
\times \nabla_{\bm{\alpha}}
\log p(\bm{\widetilde{A}}|\bm{\alpha})
\end{array}
\end{equation}
where $\beta_1$ and $\beta_2$ are two hyper-parameters to control the weights of each part. Because the reward function of policy gradient is scalar, the information from predictor and FLOPs can be easily included when updating the architecture probabilities.

Inspired by \cite{rennie2017scst}, we also add a baseline item in the reward function from the architecture obtained by greedy decoding. Specifically, we apply an $\arg\max$ operation on the probability distribution: $\hat{\bm{A}} = \mathop{\arg\max}\limits_{\bm{A}}p(\mathbf{A}|\bm{\alpha})$ to get the architecture $\hat{\bm{A}}$ and weights $\bm{\hat{\omega}}$. The baseline item is:
\begin{equation}
\hat{r} = 
\log p(\mathbf{y}|\mathbf{X},\bm{\hat{A}},\bm{\hat{\omega}}) + \beta_1 FSP(\bm{\hat{A}}) - \beta_2 FLOPs(\bm{\hat{A}}, \bm{X})
\end{equation}

Thus, the final optimization equation of self-critical training is:
\begin{equation}
\nabla_{\bm{\alpha}}L = -[r-\hat{r}]\nabla_{\bm{\alpha}}\log p(\bm{\widetilde{A}}|\bm{\alpha})
\end{equation}

At initialization, the probability distribution is set to uniform distribution for fair search. During training, the probability distribution $\bm{\alpha}$ is optimized to prefer architectures with higher accuracy from both SuperNet and the predictor, and lower FLOPs. Once trained, the final architecture is discretized by $\bm{A}_{final} = \mathop{\arg\max}\limits_{\bm{A}}p(\mathbf{A}|\bm{\alpha})$. Unlike SPOS \cite{guo2019single}, our method does not require a second stage to find the optimal architecture.

\section{Experiments}
\label{sec:intro}

\begin{table*}
	\renewcommand\arraystretch{1.2}
	\setlength\tabcolsep{3mm}
	\centering
	
	\normalsize
	\resizebox{\textwidth}{!}{
		\begin{tabular}{lcccccccccc}
			\hline
			\multirow{3}*{Methods} & \multicolumn{10}{c}{NATS-Bench TSS search space}  \\
			\cline{2-11}
			~ & \multicolumn{5}{c}{CIFAR-10} & \multicolumn{5}{c}{CIFAR-100} \\
			\cmidrule(r){2-6} \cmidrule(r){7-11} 
			
			~  & Kendall-Tau & MSE & Best Rank & Best Acc. & Cost (GPU-day) & Kendall-Tau & MSE & Best Rank & Best Acc. & Cost (GPU-day) \\
			\hline
			SPOS \cite{guo2019single} & 0.614 & 0.159 & 5301 & 91.9 & 0.11 & 0.673 & 0.302 & 4694 & 67.51 & 0.12\\
			SPOS + landmark \cite{yu2021landmark} & 0.632 & 0.153 & 4801 & 92.02 & 0.124 & 0.713 & 0.318 & 4694 & 67.51 & 0.125 \\
			SPOS + landmark + PS & 0.666 & 0.157 & 559 & 93.38 & 0.27 & 0.728 & 0.317 & 1324 & 69.63 & 0.26 \\
			SPOS + landmark + PS + predictor & 0.698 & 0.140 & 167 & 93.76 & 0.29 & 0.730 & 0.312 & 163 & 71.11 & 0.69 \\
			SPOS + landmark + PS + predictor + WWS & 0.701 & 0.135 & 94 & 93.86 & 0.33 & 0.736 & 0.272 & 112 & 71.41 & 0.74 \\

			\hline
			\hline
			\multirow{3}*{Methods} & \multicolumn{10}{c}{NATS-Bench SSS search space}  \\
			\cline{2-11}
			~ & \multicolumn{5}{c}{CIFAR-10} & \multicolumn{5}{c}{CIFAR-100} \\
			\cmidrule(r){2-6} \cmidrule(r){7-11} 
			
			~  & Kendall-Tau & MSE & Best Rank & Best Acc. & Cost (GPU-day) & Kendall-Tau & MSE & Best Rank & Best Acc. & Cost (GPU-day) \\
			\hline
			SPOS \cite{guo2019single} & 0.576 & 0.089 & 6529 & 92.13 & 0.10 & 0.702 & 0.360 & 1523 & 68.94 & 0.10\\
			SPOS + landmark \cite{yu2021landmark} & 0.579 & 0.093 & 3352 & 92.45 & 0.13 & 0.718 & 0.327 & 850 & 69.30 & 0.12 \\
			SPOS + landmark + PS & 0.587 & 0.099 & 1805 & 92.67 & 0.20 & 0.727 & 0.328 & 622 & 69.45 & 0.20 \\
			SPOS + landmark + PS + predictor & 0.623 & 0.117 & 37 & 93.34 & 0.22 & 0.727 & 0.345 & 18 & 70.63 & 0.62 \\
			SPOS + landmark + PS + predictor + WWS & 0.634 & 0.099 & 22 & 93.40 & 0.25 & 0.728 & 0.331 & 3 & 71.03 & 0.66 \\
			\hline
		\end{tabular}
	}
	\caption{Search results on NATS-Bench TSS and SSS search space. Cost is tested on a TITAN-RTX GPU. PS represents probabilistic sampling, WWS represents weak weight sharing. }
	\label{cifar}
	\vspace{-1mm}
\end{table*}

\begin{table}
	\renewcommand\arraystretch{1.2}
	\setlength\tabcolsep{3mm}
	\centering
	
	\normalsize
	\resizebox{0.5\textwidth}{!}{
		\begin{tabular}{lcccccc}
			\hline
			\multirow{3}*{Methods} & \multicolumn{6}{c}{NATS-Bench TSS search space}  \\
			\cline{2-7}
			~ & \multicolumn{3}{c}{CIFAR-10} & \multicolumn{3}{c}{CIFAR-100} \\
			\cmidrule(r){2-4} \cmidrule(r){5-7} 
			
			~  & Kendall-Tau & Corr & MSE & Kendall-Tau & Corr & MSE \\
			\hline
			MLP & 0.550 & 0.595 & 0.858 & 0.582 & 0.546 & 0.694 \\
			few-shot MLP & 0.561 & 0.539 & 0.083 & 0.617 & 0.561 & 0.132  \\
			\hline
			RNN & 0.517 & 0.493 & 0.885 & 0.524 & 0.492 & 0.707 \\
			few-shot RNN & 0.531 & 0.502 & 0.078 & 0.562 & 0.542 & 0.132  \\
			\hline
			Transformer & 0.582 & 0.609 & 1.424 & 0.587 & 0.564 & 0.407 \\
			few-shot Transformer & 0.588 & 0.601 & 0.097 & 0.595 & 0.538 & 0.122  \\

			\hline
			\hline
			\multirow{3}*{Methods} & \multicolumn{6}{c}{NATS-Bench SSS search space}  \\
			\cline{2-7}
			~ & \multicolumn{3}{c}{CIFAR-10} & \multicolumn{3}{c}{CIFAR-100} \\
			\cmidrule(r){2-4} \cmidrule(r){5-7} 
			
			~  & Kendall-Tau & Corr & MSE & Kendall-Tau & Corr & MSE \\
			\hline
			MLP & 0.795 & 0.934 & 0.687 & 0.630 & 0.759 & 0.630 \\
			few-shot MLP & 0.804 & 0.926 & 0.007 & 0.637 & 0.669 & 0.046  \\
			\hline
			RNN & 0.773 & 0.928 & 0.942 & 0.790 & 0.863 & 0.643 \\
			few-shot RNN & 0.786 & 0.861 & 0.006 & 0.812 & 0.844 & 0.048  \\
			\hline
			Transformer & 0.790 & 0.937 & 0.486 & 0.849 & 0.852 & 1.161 \\
			few-shot Transformer & 0.795 & 0.855 & 0.117 & 0.834 & 0.826 & 0.092  \\
			\hline
		\end{tabular}
	}
	\caption{Comparison between predictors trained by supervised learning and few-shot learning. MLP, RNN and Transformer based predictors are tested.}
	\label{cifar}
	\vspace{-1mm}
\end{table}

\begin{table*}
	\renewcommand\arraystretch{1.2}
	\setlength\tabcolsep{3mm}
	\centering
	
	\normalsize
	\huge
	\resizebox{\textwidth}{!}{
		\begin{tabular}{lccccccccc}
			\hline
			\multirow{2}*{Methods} & \multicolumn{3}{c}{CIFAR-10} &  \multicolumn{5}{c}{ImageNet} & \multirow{2}*{Search Method} \\
			\cmidrule(r){2-4} \cmidrule(r){5-9} 
			~ & Accuracy (\%)  & Params (M) & Search Cost (GPU-day)  & \multicolumn{2}{c}{Accuracy(\%)}  & Params (M) & FLOPs (M) & Search Cost (GPU-day) & ~\\
			\cmidrule(r){5-6}
			~ & ~  & ~ & ~  & top-1 & top-5  & ~ & ~ & ~ & ~\\

			\hline
			DenseNet-BC \cite{huang2016densely} & 96.54 & 25.6 & - & - & - & - & - & - & manual\\
			Inception-v1 \cite{Szegedy2015} & - & - & - & 69.8 & 89.9 & 6.6 & 1448  & - & manual \\
			MobileNet \cite{howard2017mobilenets} & - & - & - & 70.6 & 89.5 & 4.2 & 569  & - & manual\\
			ShuffleNet 2$\times$ (v2) \cite{ma2018shufflenet} & - & - & - & 74.9 & - & $\sim$5 & 591 & - & manual\\
			\hline
			NASNet-A + cutout \cite{Zoph_2018_CVPR} & 97.35 & 3.3  & 1800 & 74.0 & 91.6 & 5.3 & 564 & 1800  & RL\\
			AmoebaNet +cutout \cite{real2019regularized} & 97.45$\pm$0.05 & 2.8 & 3150 & 75.7 & 92.4 & 6.4 & 570 & 3150  & evolution \\
			PNAS \cite{liu2018pnas} & 96.59$\pm$0.09 & 3.2 & 225 & 74.2 & 91.2 & 5.1 & 588 & 225  & SMBO\\
			ENAS + cutout \cite{pham2018efficient} & 97.11 & 4.6 & 0.5 & - & - & - & - & - & RL\\

			
			\hline
			DARTS (1st order) + cutout \cite{liu2018darts} & 97.00$\pm$0.14 & 3.3  & 0.4 & - & - & - & - & - & gradient \\
			DARTS (2nd order) + cutout \cite{liu2018darts} & 97.24$\pm$0.09 & 3.3 & 4.0 & 73.2 & 91.3 & 4.7 & 574 & 4.0 & gradient \\
			PARSEC + cutout \cite{casale2019PARSEC} & 97.19$\pm$0.03 & 3.7 & 1.0 & 74.0 & 91.6 & 5.6 & - & 1.0 & gradient \\
			SNAS (moderate) + cutout \cite{xie2018snas} & 97.02$\pm$0.02 & 2.8 & 1.5 & 72.7 & 90.8 & 4.3 & 522 & 1.5 & gradient \\
			P-DARTS+cutout \cite{chen2019progressive} & 97.5 & 3.4 & 0.3 & 75.6 & 92.6 & 4.9 & 557 & 0.3 & gradient\\
			NASP + cutout \cite{yao2020efficient} & 97.17$\pm$0.09 & 3.3 & 0.1 & - & - & - & - & - & gradient\\
			PC-DARTS + cutout \cite{xu2019pc} & 97.43$\pm$0.07 & 3.6 & 0.1 & 74.9 & 92.2 & 5.3 & 597 & 3.8 &  gradient\\
			FairDARTS + cutout \cite{chu2020fairdarts} & 97.46$\pm$0.05 & 3.3 & 0.1 & 75.6 & 92.6 & 4.3 & 440 & 3 &  gradient\\ 			
			ISTA-NAS + cutout \cite{yang2020istanas} & 97.64$\pm$0.06 & 3.37 & 2.3& 76.0 & 92.9 & 5.65 & 638 & 4.2 & gradient\\
			DOTS + cutout \cite{gu2021dots} & 97.51$\pm$0.06 & 3.5 & 0.26 & 76.0 & 92.8 & 5.3 & 596 & 1.3 & gradient\\
			
			IDARTS + cutout \cite{xue2021idarts} & 97.68 & 4.16 & 0.1 & 76.52 & 93.0 & 6.18 & 714 & 3.8 & gradient\\
			VIM-NAS + cutout \cite{wang2021vimnas} & 97.55$\pm$0.04 & 3.9 & 0.007 & 76.2 & 92.9 & - & 660 & 0.26 & gradient\\
			\hline
			WPNAS-A + cutout & 97.45$\pm$0.10 & 2.4 & 1.5 & 76.22 & 92.70 & 5.03 & 550 & 1.5 & probabilistic sample\\
			WPNAS-B + cutout & 97.70$\pm$0.05 & 4.8 & 2.0 & 76.61 & 92.98 & 7.56 & 848 & 2.0 & probabilistic sample\\
			WPNAS-A + cutout$^\dagger$ & 98.00$\pm$0.12 & 2.5 & 1.5 & 76.65 & 93.01 & 5.30 & 553 & 1.5 & probabilistic sample\\
			WPNAS-B + cutout$^\dagger$ & 98.12$\pm$0.08 & 5.2 & 2.0 & 76.81 & 93.16 & 7.90 & 852 & 2.0 & probabilistic sample\\
			\hline
		\end{tabular}
	}
	\caption{Search results on CIFAR-10 and ImageNet under DARTS search space and comparison with state-of-the-art methods. Cost is tested on a TITAN-RTX GPU. $^\dagger$ represents using Swish, SE and Autoaugment.}
	\label{cifar}
	\vspace{-1mm}
\end{table*}

\begin{table*}
	\renewcommand\arraystretch{1.2}
	\setlength\tabcolsep{3mm}
	\centering
	
	\normalsize
	\resizebox{\textwidth}{!}{
		\begin{tabular}{lccccccccc}
			\hline
			\multirow{2}*{Methods} & \multicolumn{3}{c}{CIFAR-10} &  \multicolumn{5}{c}{ImageNet} & \multirow{2}*{Search Method} \\
			\cmidrule(r){2-4} \cmidrule(r){5-9} 
			~ & Accuracy (\%)  & Params (M) & Search Cost (GPU-day)  & \multicolumn{2}{c}{Accuracy(\%)}  & Params (M) & FLOPs (M) & Search Cost (GPU-day) & ~\\
			\cmidrule(r){5-6}
			~ & ~  & ~ & ~  & top-1 & top-5  & ~ & ~ & ~ & ~\\

			\hline
			DenseNet-BC \cite{huang2016densely} & 96.54 & 25.6 & - & - & - & - & - & - & manual\\
			Inception-v1 \cite{Szegedy2015} & - & - & - & 69.8 & 89.9 & 6.6 & 1448  & - & manual \\
			ShuffleNet 2$\times$ (v2) \cite{ma2018shufflenet} & - & - & - & 74.9 & - & $\sim$5 & 591 & - & manual\\
			MobileNet \cite{howard2017mobilenets} & - & - & - & 70.6 & 89.5 & 4.2 & 569  & - & manual\\
			MobileNetV2 \cite{sandler2018mobilenetv2} & - & - & - & 72.0 & - & 3.4 & 300  & - & manual\\
			MobileNetV3 \cite{howard2019mobilenetv3} & - & - & - & 75.2 & - & 5.4 & 219  & - & manual\\

			
			\hline
			SPOS \cite{guo2019single} & - & - & - & 74.7 & - & - & 328 & 29.0 & uniform sample \\
			FBNet \cite{wu2019fbnet} & - & - & - & 74.9 & - & 5.5 & 375 & 9.0 & gradient \\
			FBNetV2 \cite{wan2020fbnetv2} & - & - & - & 77.2 & - & - & 325 & 0.6k & gradient \\
			FairNAS \cite{chu2021fairnas} & - & - & - & 75.3 & - & 4.6 & 388 & 12.0 & uniform sample \\
			FairNAS$^\dagger$ \cite{chu2021fairnas} & 98.2 (224*224 resolution) & - & - & 77.5 & - & 5.9 & 392 & 12.0 & uniform sample \\
			FP-NAS + AutoAugment \cite{yan2021fpnas} & - & - & - & 76.6 & - & 6.4 & 268 & 7.6 & probabilistic sample \\
			MCT-NAS-C$^\diamond$ \cite{su2021mctnas} & - & - & - & 76.3 & 92.6 & 4.9 & 280 & 1 & prioritized sample\\
			
			\hline
			WPNAS-A  & 95.4$\pm$0.06 & 3.4 & 1.0 & 75.6 & 92.4 & 4.7 & 390 & 1.0 & probabilistic sample\\
			WPNAS-B  & 96.06$\pm$0.04 & 5.7 & 1.3 & 77.0 & 93.4 & 6.9 & 492 & 1.3 & probabilistic sample\\
			WPNAS-A + AutoAugment & 96.4$\pm$0.10 & 3.4 & 1.0 & 76.4 & 92.8 & 4.7 & 390 & 1.0 & probabilistic sample\\
			WPNAS-B + AutoAugment & 96.52$\pm$0.12 & 5.7 & 1.3 & 77.8 & 93.6 & 6.9 & 492 & 1.3 & probabilistic sample\\
			\hline
		\end{tabular}
	}
	\caption{Search results on CIFAR-10 and ImageNet under MobileNet search space and comparison with state-of-the-art methods. $^\dagger$ represents using Swish, SE and Autoaugment. $^\diamond$ represents using training tricks including WarmUp and EMA \cite{he2020ema}.}
	\label{cifar}
	\vspace{-1mm}
\end{table*}

To validate the advantages of the proposed algorithm, we conduct experiments on three commonly used search spaces: NATS-Bench search space \cite{dong2020natsbench} (section 4.1), DARTS search space \cite{liu2018darts} (section 4.2) and MobileNet search space \cite{sandler2018mobilenetv2} (section 4.3). 

\noindent\textbf{WPNAS procedure}. The whole process of our WP-NAS algorithm contains 5 steps. 1) \textbf{WarmUp}. We train the SuperNet without optimizing the architecture probabilities for 200 epochs. Only weights are updated in this step. We apply SGD optimizer with initial learning rate 0.1 and follow a cosine schedule to decay the learning rate to 0.01.  2) \textbf{Acquisition of the ground truth of the Predictor}. Uniform sampling 200 architectures, inheriting weights from the SuperNet and train these architectures for 30 epochs. Thus, we can obtain 200 architecture-accuracy data pairs as the ground truth of the predictor. We use the proposed way to get the ground truth instead of training from scratch for the sake of saving time. The Kendall Tau between our GT and that obtained by training from scratch is: Kendall Tau = 0.7, but the training time reduced by 85\%. 3) \textbf{Train the predictor}. We use the proposed few-shot learning based algorithm to train the predictor. We split the 200 GTs with 170 architectures for training and 30 architectures for validation. The support size is set to 30. We train the predictor for 100 epochs. 4) \textbf{WPNAS Search}. Alternatively updating the weights and architecture probabilities for 300 epochs. Architecture probabilities are updated based on equation (7) by self-critical policy gradient training with a combined reward of direct evaluation, output of the predictor and the FLOPs. We also apply an SGD optimizer with cosine decay scheduler from 0.01 to 1e-5. 5) \textbf{Evaluation}. Obtain the final searched architecture with an argmax operation and train this architecture from scratch for 600 epochs.

\begin{figure*}[t]
	
	\begin{subfigure}{0.45\textwidth}
		\centering
		\includegraphics[width=1\textwidth,height=0.33\textwidth]{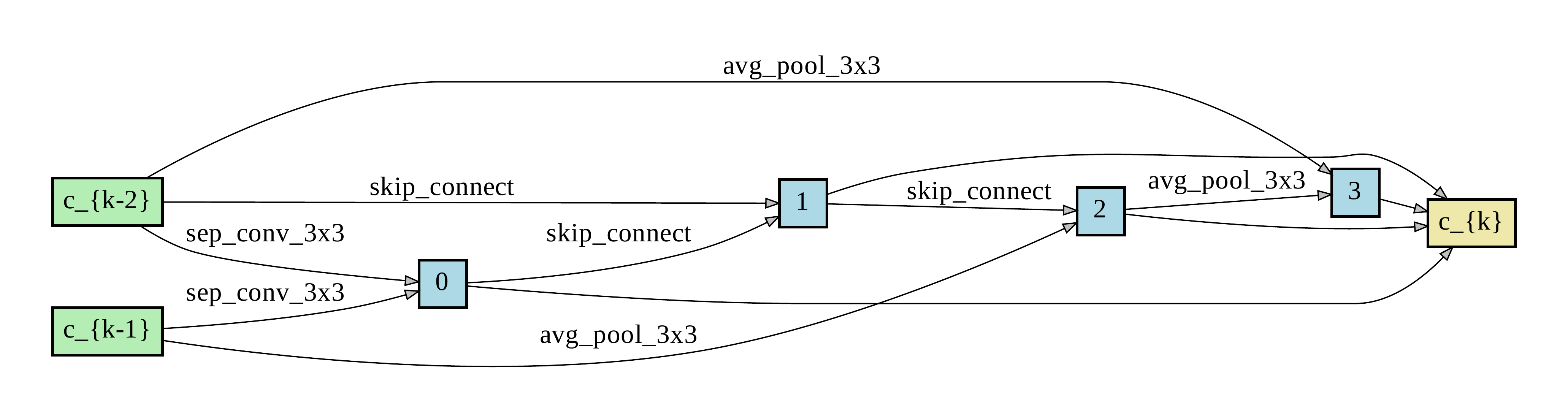}
		\caption{WPNAS-A, normal cell}
	\end{subfigure}
	\hspace{2mm}
	\vspace{10mm}
	\begin{subfigure}{0.55\textwidth}
		\centering
		\includegraphics[width=0.9\textwidth,height=0.33\textwidth]{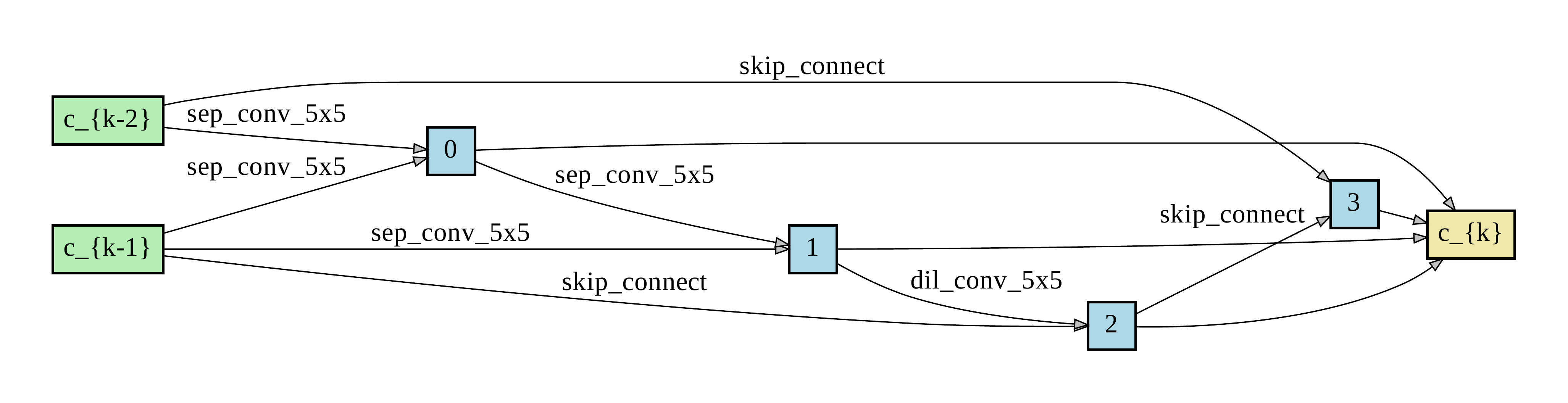}
		\caption{WPNAS-A, reduction cell}
	\end{subfigure}
	
	\begin{subfigure}{0.5\textwidth}
		\centering
		\includegraphics[width=0.9\textwidth,height=0.34\textwidth]{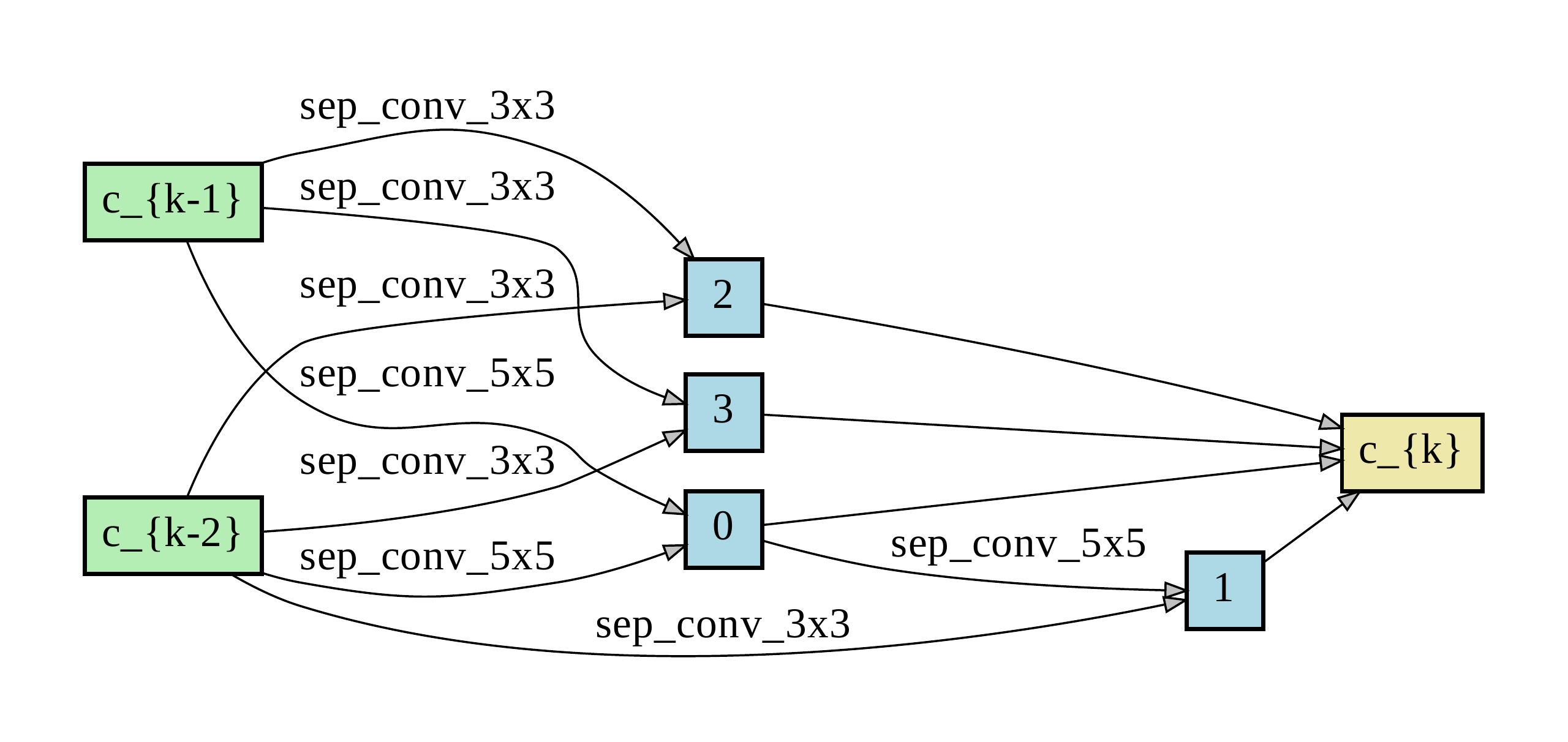}
		\caption{WPNAS-B, normal cell}
	\end{subfigure}
	\hspace{2mm}
	\begin{subfigure}{0.5\textwidth}
		\centering
		\includegraphics[width=0.9\textwidth,height=0.34\textwidth]{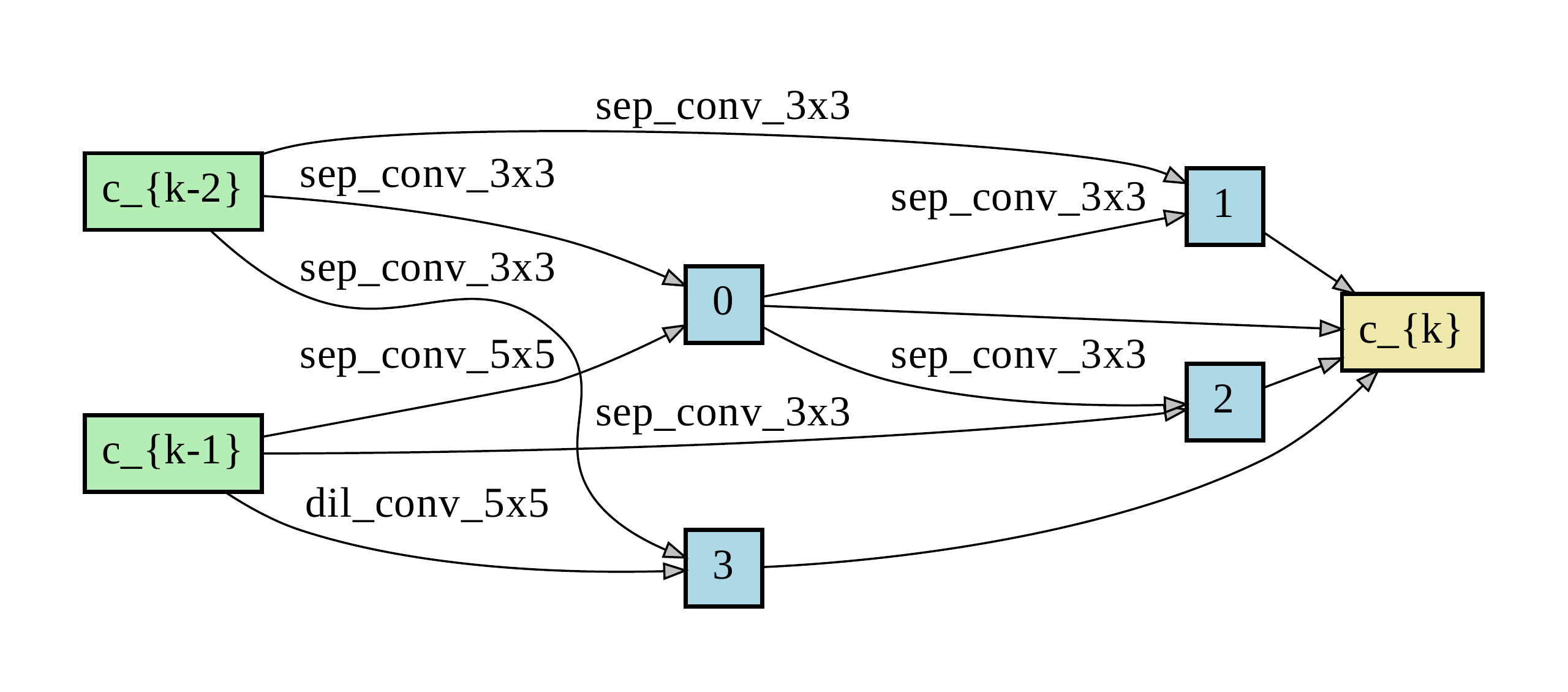}
		\caption{WPNAS-B, reduction cell}
	\end{subfigure}
	
	\caption{Searched architectures of WPNAS-A ((a) and (b)) and WPNAS-B ((c) and (d)).}
	\label{supp_f3}
\end{figure*}

\subsection{NATS-Bench experiments}

\noindent\textbf{NATS-Bench}. NATS-Bench \cite{dong2020natsbench} is an extension version of NAS-Bench-201 \cite{dong2020bench201} and it is a unified benchmark on searching for both topology and size. For topology search space (TSS), it is a cell-based search space inspired by some cell-based algorithms \cite{liu2018darts, pham2018efficient}. A cell is represented as a densely-connected directed acyclic graph (DAG) and all cells have the same topology. The DAG contains 4 nodes and 6 edges, and each edge of the DAG has 5 candidate operations. Thus, there are $5^6=15625$ different architectures in the TSS search space. For size search space (SSS), the cell is the best one in the TSS search space on CIFAR-100 dataset. Each architecture contains five layers with a unique configuration based on the number of channels in each layer. There are 8 candidate channel numbers for searching: $\{8,16,24,32,40,48,56,64\}$. Therefore, the search space size are $8^5=32768$. We perform experiments on CIFAR-10 and CIFAR-100 datasets \cite{krizhevsky2009cifar}.

\noindent\textbf{Results on NATS-Bench}. Table 1. shows the results on NATS-Bench. \textbf{Baselines}. We apply single-path one-shot (SPOS) \cite{guo2019single}, the most representative sampling based algorithm as the baseline. We train the SuperNet for 500 epochs with uniform sampling. Similar as \cite{guo2019single}, we use an evolutional algorithm to find the optimal architecture. Another baseline is adding landmark regularization \cite{yu2021landmark} when training the SuperNet. Since we have several GTs of the predictor, we can make use of these data as the landmark to better guide the ranking between architectures during SuperNet training. We also train the SuperNet for 500 epochs with landmark regularization and also apply a second stage to find the optimal architecture. \textbf{Metrics}. We report the ranking correlation in terms of two metrics: Kendall-Tau and mean square error (MSE). We also report the rank and accuracy of the searched architecture and the overall cost. 

We gradually add probability sampling, few-shot predictor and weakly weight sharing mechanism, and compare the results with baselines. Experiments on CIFAR-10 and CIFAR-100 under TSS and SSS search space have consistent results, that is, with the introduction of the proposed novel methods, the ranking performance is gradually improved, and the ranking and accuracy of the final architecture are also improved accordingly. It should be noted that after the introduction of each method, the cost will increase accordingly, especially after the introduction of predictor. However, compared with baseline, the cost is also in the same order of magnitude and less than 1 GPU-day, which is also acceptable.

\noindent\textbf{Few-shot Predictor}. Table 2. shows the results of the proposed few-shot predictor. The baseline is supervised training whose input is a single architecture and the output is a scalar (validation accuracy). We test three types of the predictor, which are MLP, RNN and Transformer based. For both supervised learning and few-shot learning, the training / validation set are both 170/30. We train the predictor for 100 epochs using SGD with learning rate 1e-4. We apply a ranking loss similar as \cite{xu2021renas}. We report three metrics: Kendall-Tau, Pearson Correlation Coefficient and Mean Square Error (MSE). The results show that few-shot predictor has consistent better Kendall-Tau and MSE on CIFAR-10 and CIFAR-100 under TSS and SSS search spaces. The MSEs of few-shot predictor are extremely lower than that of supervised learning, because the proposed few-shot predictor can better learn the relationships between architectures, even only ranking loss is used without any other regression guidance.

\subsection{DARTS search space}
To compare with other NAS algorithms, we also conduct experiments on the commonly used DARTS search space \cite{liu2018darts} on CIFAR-10 and ImageNet \cite{olga2015imagenet} datasets. DARTS search space is also a cell based search space. The DAG contains 4 inner nodes, and each node has two parents nodes. The search space size is $3.3\times10^{13}$. We set two types of architecture parameters for each node of the DAG: operation selection and parent node selection. Operation selection is used to select one operation from 8 candidates. Parent node selection is designed to select two parents nodes from the current node's all previous nodes. Suppose the current node has $N$ previous nodes, the number of candidate combinations are $\mathrm{C}_N^2$. Suppose operation selection and parent node selection are independent to each other, equation (1) remains valid as well.  All the rest experiment settings are identical to DARTS for fair comparison. We compare our method with several manual networks such as DenseNet \cite{huang2016densely} and MobileNet \cite{howard2017mobilenets} and some SOTA NAS algorithms on DARTS search space \cite{liu2018darts, chen2019progressive, chu2020fairdarts}. Similar as DARTS, we conduct the search experiment on CIFAR-10 dataset and validate the performance of the searched architecture on CIFAR-10 and ImageNet dataset. The channel number is expanded on ImageNet. Similar as \cite{liu2018darts, chu2020fairdarts}, we also run 3 times evaluation on CIFAR-10 to get a more accurate result.

Table 3. shows the results on DARTS search space. WPNAS-A and WPNAS-B are two searched architectures with different $\beta_2$ (WPNAS-A: $\beta_2=1^{-3}$, WPNAS-B: $\beta_2=1^{-5}$). Figure 4. shows the searched architectures of WPNAS-A and WPNAS-B. Larger $\beta_2$ can lead to find smaller architectures, the accuracy of WPNAS-A on CIFAR-10 dataset is 97.45 which is comparable with other SOTA methods, but with a smaller model size (2.4M). The larger architecture WPNAS-B has better accuracy 97.70 which is better than other methods. The results of our method on ImageNet (WPNAS-A:76.22, WPNAS-B:76.61) is comparable with other methods with comparable Params and FLOPs. Similar with \cite{chu2020fairdarts, chu2021fairnas}, we also add some tricks such as Swish, SE and AutoAugment, the performances of WPNAS-A and WPNAS-B are all improved significantly and achieves SOTA performance. 

\subsection{MobileNet search space}
MobileNet search space is another commonly used macro search space for NAS benchmarking. Following \cite{su2021mctnas}, we construct the MobileNet-like SuperNet with 17 search blocks, and each block is a MobileNetV2 inverted bottleneck \cite{sandler2018mobilenetv2}. For each search block, the candidate set of convolutional kernel size is \{3,5\}, the expansion ratio candidate set is \{1,3,6\}, nonlinearities is selected in \{ReLU, HardSwish \cite{howard2019mobilenetv3}\} and each block can choose to use Squeeze-and-Excitation (SE) \cite{hu2018senet} module or not. The search space size is $24^{17}\approx2.9\times 10^{23}$. We compare our NAS algorithm with some SOTA algorithms on this search space including SPOS \cite{guo2019single}, FBNet \cite{wu2019fbnet, wan2020fbnetv2}, FairNas \cite{chu2021fairnas}, FP-NAS \cite{yan2021fpnas} and MCT-NAS \cite{su2021mctnas}. We also conduct the search experiment on CIFAR-10 dataset and transfer the architecture on ImageNet. The channel number is expanded on ImageNet.

Table 4. shows the results on MobileNet search space.  We also obtain two final architectures WPNAS-A and WPNAS-B (WPNAS-A: $\beta_2=1^{-3}$, WPNAS-B: $\beta_2=1^{-5}$). On ImageNet dataset, WPNAS-A has higher top-1 accuracy (75.6) compared with other methods (SPOS, FBNet and FairNAS) with similar Params and FLOPs. WPNAS-A is a bit lower than MCT-NAS-C, but MCT-NAS use other training tricks including WarmUp and EMA \cite{he2020ema}. WPNAS-B also has higher accuracy than FP-NAS with similar Params. FBNetV2 has higher accuracy, but its cost (0.6k) is also much higher than ous. With the help of AutoAugment, WPNAS-B (77.8) also surpass FairNAS $^\dagger$ (77.5).

\section{Conclusion}
\label{sec:intro}

In this paper, we propose to jointly use weight sharing and predictor in a unified framework. Architecture coupling caused by weight sharing can lead to big evaluation ranking gap between architectures using weight sharing training and stand-alone training. To increase the correctness of the evaluation of architectures, besides the direct evaluation using the weights from SuperNet, we also introduce a predictor to evaluate the architecture from another perspective. The evaluations from these two parts are combined as the final prediction of the architecture, and are used as the reward of a self-critical policy gradient algorithm. After the introduction of predictor, the evaluation of architecture will be more accurate, which can also better guide the NAS algorithm to find better architectures. Besides, we propose a novel few-shot learning based predictor to enhance the performance of the predictor. Furthermore, we propose a weakly weight sharing strategy by introducing a HyperNet to reduce the degree of weight sharing, thus leading to more stand-alone-training like SuperNet training. We conduct comprehensive experiments on CIFAR and ImageNet datasets under NATS-Bench, DARTS and MobileNet search space and our algorithm achieves SOTA performances on these datasets.

{\small
\bibliographystyle{ieee_fullname}
\bibliography{egbib}

\begin{thebibliography}{10}\itemsep=-1pt

\bibitem{casale2019PARSEC}
Francesco~Paolo Casale, Jonathan Gordon, and Nicolo' Fusi.
\newblock Probabilistic neural architecture search.
\newblock In {\em arXiv:1902.05116}, 2019.

\bibitem{chang2019data}
Jianlong Chang, Yiwen Guo, GAOFENG Meng, SHIMING Xiang, Chunhong Pan, et~al.
\newblock Data: Differentiable architecture approximation.
\newblock In {\em NeurIPS}, pages 874--884, 2019.

\bibitem{chen2019progressive}
Xin Chen, Lingxi Xie, Jun Wu, and Qi Tian.
\newblock Progressive differentiable architecture search: Bridging the depth
  gap between search and evaluation.
\newblock In {\em ICCV}, pages 1294--1303, 2019.

\bibitem{chen2021comparator}
Yaofo Chen, Yong Guo, Qi Chen, Minli Li, Wei Zeng, Yaowei Wang, and Mingkui
  Tan.
\newblock Contrastive neural architecture search with neural architecture
  comparators.
\newblock In {\em CVPR}, 2021.

\bibitem{chen2020adc}
Zewei Chen, Fengwei Zhou, George Trimponias, and Li Zhenguo.
\newblock Multi-objective neural architecture search via non-stationary policy
  gradient.
\newblock In {\em arXiv:2001.08437}, 2020.

\bibitem{chu2021darts-}
Xiangxiang Chu, Xiaoxing Wang, Zhang Bo, Shun Lu, Xiaolin Wei, and Junchi Yan.
\newblock Darts-: robustly stepping out of performance collapse without
  indicators.
\newblock In {\em ICLR}, 2021.

\bibitem{chu2021fairnas}
Xiangxiang Chu, Bo Zhang, and Xu Ruijun.
\newblock Fairnas: Rethinking evaluation fairness of weight sharing neural
  architecture search.
\newblock In {\em ICCV}, 2021.

\bibitem{chu2020fairdarts}
Xiangxiang Chu, Tianbao Zhou, Bo Zhang, and Jixiang Li.
\newblock Fair darts: Eliminating unfair advantages in differentiable
  architecture search.
\newblock In {\em ECCV}, 2020.

\bibitem{dong2020natsbench}
Xuanyi Dong, Lu Liu, Katarzyna Musial, and Bogdan Gabrys.
\newblock Nats-bench: Benchmarking nas algorithms for architecture topology and
  size.
\newblock In {\em IEEE transactions on pattern analysis and machine
  intelligence}, 2021.

\bibitem{dong2019searching}
Xuanyi Dong and Yi Yang.
\newblock Searching for a robust neural architecture in four gpu hours.
\newblock In {\em CVPR}, pages 1761--1770, 2019.

\bibitem{dong2020bench201}
Xuanyi Dong and Yi Yang.
\newblock Nas-bench-201: Extending the scope of reproducible neural
  architecture search.
\newblock In {\em ICLR}, 2020.

\bibitem{dudziak2020gcn}
Łukasz Dudziak, Thomas Chau, Mohamed~S. Abdelfattah, Royson Lee, Hyeji Kim,
  and Nicholas~D. Lane.
\newblock Brp-nas: Prediction-based nas using gcns.
\newblock In {\em arXiv:2007.08668}, 2020.

\bibitem{gu2021dots}
Yu-Chao Gu, Li-Juan Wang, Yun Liu, Yi Yang, Yu-Huan Wu, Shao-Ping Lu, and
  Ming-Ming Cheng.
\newblock Dots: Decoupling operation and topology in differentiable
  architecture search.
\newblock In {\em CVPR}, 2021.

\bibitem{guo2019single}
Zichao Guo, Xiangyu Zhang, Haoyuan Mu, Wen Heng, Zechun Liu, Yichen Wei, and
  Jian Sun.
\newblock Single path one-shot neural architecture search with uniform
  sampling.
\newblock {\em arXiv preprint arXiv:1904.00420}, 2019.

\bibitem{he2020ema}
Kaiming He, Haoqi Fan, Yuxin Wu, Saining Xie, and Ross Girshick.
\newblock Momentum contrast for unsupervised visual representation learning.
\newblock In {\em CVPR}, 2020.

\bibitem{howard2019mobilenetv3}
Andrew Howard, Mark Sandler, Grace Chu, Liang-Chieh Chen, Bo Chen, Mingxing
  Tan, Weijun Wang, Yukun Zhu, Ruoming Pang, Vijay Vasudevan, Quoc~V. Le, and
  Hartwig Adam.
\newblock Searching for mobilenetv3.
\newblock In {\em ICCV}, 2019.

\bibitem{howard2017mobilenets}
Andrew~G Howard, Menglong Zhu, Bo Chen, Dmitry Kalenichenko, Weijun Wang,
  Tobias Weyand, Marco Andreetto, and Hartwig Adam.
\newblock Mobilenets: Efficient convolutional neural networks for mobile vision
  applications.
\newblock {\em arXiv preprint arXiv:1704.04861}, 2017.

\bibitem{hu2018senet}
Jie Hu, Li Shen, and Gang Sun.
\newblock Squeeze-and-excitation networks.
\newblock In {\em CVPR}, 2018.

\bibitem{huang2016densely}
G. Huang, Z. Liu, L. van~der Maaten, and K. Weinberger.
\newblock Densely connected convolutional networks.
\newblock In {\em CVPR}, 2017.

\bibitem{huang2021generator}
Sian-Yao Huang and Wei-Ta Chu.
\newblock Searching by generating: Flexible and efficient one-shot nas with
  architecture generator.
\newblock In {\em CVPR}, 2021.

\bibitem{krizhevsky2009cifar}
Alex Krizhevsky and Geoffrey Hinton.
\newblock Learning multiple layers of features from tiny images.
\newblock In {\em Citeseer, Tech. Rep}, 2009.

\bibitem{liu2018pnas}
Chenxi Liu, Barret Zoph, Maxim Neumann, Jonathon Shlens, Wei Hua, Li-Jia Li,
  Fei-Fei Li, Alan Yuille, Jonathan Huang, and Kevin Murphy.
\newblock Progressive neural architecture search.
\newblock In {\em ECCV}, pages 19--34, 2018.

\bibitem{liu2018darts}
Hanxiao Liu, Karen Simonyan, and Yiming Yang.
\newblock Darts: Differentiable architecture search.
\newblock In {\em ICLR}, 2019.

\bibitem{ma2018shufflenet}
Ningning Ma, Xiangyu Zhang, Hai-Tao Zheng, and Jian Sun.
\newblock Shufflenet v2: Practical guidelines for efficient cnn architecture
  design.
\newblock In {\em ECCV}, pages 116--131, 2018.

\bibitem{ning2020gates}
Xuefei Ning, Yin Zheng, Tianchen Zhao, Yu Wang, and Huazhong Yang.
\newblock A generic graph-based neural architecture encoding scheme for
  predictor-based nas.
\newblock In {\em ECCV}, 2020.

\bibitem{pham2018efficient}
Hieu Pham, Melody~Y Guan, Barret Zoph, Quoc~V Le, and Jeff Dean.
\newblock Efficient neural architecture search via parameter sharing.
\newblock In {\em ICML}, 2018.

\bibitem{real2019regularized}
Esteban Real, Alok Aggarwal, Yanping Huang, and Quoc~V Le.
\newblock Regularized evolution for image classifier architecture search.
\newblock In {\em AAAI}, volume~33, pages 4780--4789, 2019.

\bibitem{rennie2017scst}
Steven~J. Rennie, Etienne Marcheret, Youssef Mroueh, Jarret Ross, and Vaibhava
  Goel.
\newblock Self-critical sequence training for image captioning.
\newblock In {\em CVPR}, 2017.

\bibitem{olga2015imagenet}
Olga Russakovsky, Jia Deng, Hao Su, Jonathan Krause, Sanjeev Satheesh, Sean Ma,
  Zhiheng Huang, Andrej Karpathy, Aditya Khosla, Michael Bernstein,
  Alexander~C. Berg, and Fei-Fei Li.
\newblock Imagenet large scale visual recognition challenge.
\newblock In {\em IJCV}, 2015.

\bibitem{sandler2018mobilenetv2}
Mark Sandler, Andrew Howard, Menglong Zhu, Andrey Zhmoginov, and Liang-Chieh
  Chen.
\newblock Mobilenetv2: Inverted residuals and linear bottlenecks.
\newblock In {\em CVPR}, 2018.

\bibitem{su2021mctnas}
Xiu Su, Tao Huang, Yanxi Li, Shan You, Fei Wang, Chen Qian, Changshui Zhang,
  and Chang Xu.
\newblock Prioritized architecture sampling with monto-carlo tree search.
\newblock In {\em CVPR}, 2021.

\bibitem{sung2018relation}
Flood Sung, Yongxin Yang, Li Zhang, Tao Xiang, Philip~H.S. Torr, and Timothy~M.
  Hospedales.
\newblock Learning to compare: Relation network for few-shot learning.
\newblock In {\em CVPR}, 2018.

\bibitem{Szegedy2015}
C. Szegedy, W. Liu, Y. Jia, P. Sermanet, S. Reed, D. Anguelov, D. Erhan, V.
  Vanhoucke, and A. Rabinovich.
\newblock Going deeper with convolutions.
\newblock In {\em CVPR}, 2015.

\bibitem{tan2019mnasnet}
Mingxing Tan, Bo Chen, Ruoming Pang, Vijay Vasudevan, Mark Sandler, Andrew
  Howard, and Quoc~V Le.
\newblock Mnasnet: Platform-aware neural architecture search for mobile.
\newblock In {\em CVPR}, pages 2820--2828, 2019.

\bibitem{tang2020semi}
Yehui Tang, Yunhe Wang, Yixing Xu, Hanting Chen, and Boxin Shi.
\newblock A semi-supervised assessor of neural architectures.
\newblock In {\em CVPR}, 2020.

\bibitem{wan2020fbnetv2}
Alvin Wan, Xiaoliang Dai, Peizhao Zhang, Zijian He, Yuandong Tian, Saining Xie,
  Bichen Wu, Matthew Yu, Kan Chen, Peter Vajda, and Joseph~E. Gonzalez.
\newblock Fbnetv2: Differentiable neural architecture search for spatial and
  channel dimensions.
\newblock In {\em CVPR}, 2020.

\bibitem{wang2021vimnas}
Yaoming Wang, Yuchen Liu, Wenrui Dai, Chenglin Li, Junni Zou, and Hongkai
  Xiong.
\newblock Learning latent architectural distribution in differentiable neural
  architecture search via variational information maximization.
\newblock In {\em ICCV}, 2021.

\bibitem{wen2020predictor}
Wei Wen, Hanxiao Liu, Hai Li, Yiran Chen, Gabriel Bender, and Pieter-Jan
  Kindermans.
\newblock Neural predictor for neural architecture search.
\newblock In {\em ECCV}, 2020.

\bibitem{wu2019fbnet}
Bichen Wu, Xiaoliang Dai, Peizhao Zhang, Yanghan Wang, Fei Sun, Yiming Wu,
  Yuandong Tian, Peter Vajda, Yangqing Jia, and Kurt Keutzer.
\newblock Fbnet: Hardware-aware efficient convnet design via differentiable
  neural architecture search.
\newblock In {\em CVPR}, pages 10734--10742, 2019.

\bibitem{xie2018snas}
Sirui Xie, Hehui Zheng, Chunxiao Liu, and Liang Lin.
\newblock Snas: stochastic neural architecture search.
\newblock In {\em ICLR}, 2019.

\bibitem{xu2021renas}
Yixing Xu, Yunhe Wang, Kai Han, Yehui Tang, Shangling Jui, Chunjing Xu, and
  Chang Xu.
\newblock Renas: Relativistic evaluation of neural architecture search.
\newblock In {\em CVPR}, 2021.

\bibitem{xu2019pc}
Yuhui Xu, Lingxi Xie, Xiaopeng Zhang, Xin Chen, Guo-Jun Qi, Qi Tian, and
  Hongkai Xiong.
\newblock Pc-darts: Partial channel connections for memory-efficient
  differentiable architecture search.
\newblock In {\em ICLR}, 2020.

\bibitem{xue2021idarts}
Song Xue, Runqi Wang, Baochang Zhang, Tian Wang, Guodong Guo, and David
  Doermann.
\newblock Idarts: Interactive differentiable architecture search.
\newblock In {\em ICCV}, 2021.

\bibitem{yan2021fpnas}
Zhicheng Yan, Xiaoliang Dai, Peizhao Zhang, Yuandong Tian, Bichen Wu, and Matt
  Feiszli.
\newblock Fp-nas: Fast probabilistic neural architecture search.
\newblock In {\em CVPR}, 2021.

\bibitem{yang2020istanas}
Yibo Yang, Hongyang Li, Shan You, Fei Wang, Chen Qian, and Zhouchen Lin.
\newblock Ista-nas: Efficient and consistent neural architecture search by
  sparse coding.
\newblock In {\em NeurIPS}, 2020.

\bibitem{yang2020cars}
Zhaohui Yang, Yunhe Wang, Xinghao Chen, Boxin Shi, Chao Xu, Chunjing Xu, Qi
  Tian, and Chang Xu.
\newblock Cars: Continuous evolution for efficient neural architecture search.
\newblock In {\em CVPR}, 2020.

\bibitem{yao2020efficient}
Quanming Yao, Ju Xu, Wei-Wei Tu, and Zhanxing Zhu.
\newblock Efficient neural architecture search via proximal iterations.
\newblock In {\em AAAI}, 2020.

\bibitem{ying2019bench101}
Chris Ying, Aaron Klein, Esteban Real, Eric Christiansen, Kevin Murphy, and
  Frank Hutter.
\newblock Nas-bench-101: Towards reproducible neural architecture search.
\newblock In {\em ICML}, 2019.

\bibitem{you2020greedy}
Shan You, Tao Huang, Mingming Yang, Fei Wang, Chen Qian, and Changshui Zhang.
\newblock Greedynas: Towards fast one-shot nas with greedy supernet.
\newblock In {\em CVPR}, 2020.

\bibitem{yu2021landmark}
Kaicheng Yu, Ren´e Ranftl, and Mathieu Salzmann.
\newblock Landmark regularization: Ranking guided super-net training in neural
  architecture search.
\newblock In {\em CVPR}, 2021.

\bibitem{zoph}
B. Zoph and Q. Le.
\newblock Neural architecture search with reinforcement learning.
\newblock In {\em ICLR}, 2017.

\bibitem{Zoph_2018_CVPR}
Barret Zoph, Vijay Vasudevan, Jonathon Shlens, and Quoc~V. Le.
\newblock Learning transferable architectures for scalable image recognition.
\newblock In {\em CVPR}, June 2018.

\end{thebibliography}
}

\end{document}